\documentclass[letterpaper, 10 pt, conference]{ieeeconf}  

\IEEEoverridecommandlockouts                              

\usepackage{graphicx}
\usepackage{amsmath,amssymb} 
\usepackage{color}
\usepackage{multirow}
\usepackage{tabularx}
\usepackage{comment}
\usepackage{tikz,pgfplots}

\def\eg{\emph{e.g.}} 
\def\ie{\emph{i.e.}} 
\def\etal{\emph{et al.~}}
\definecolor{brightgreen}{rgb}{0.4, 1.0, 0.0}
\definecolor{darkorchid}{rgb}{0.6, 0.2, 0.8}
\definecolor{cadmiumgreen}{rgb}{0.0, 0.42, 0.24}
\definecolor{aqua}{rgb}{0.0, 1.0, 1.0}
\definecolor{burntorange}{rgb}{0.8, 0.33, 0.0}
\definecolor{chromeyellow}{rgb}{1.0, 0.65, 0.0}

\newcommand{\figref}[1]{Fig. \ref{fig:#1}}
\newcommand{\tabref}[1]{Table \ref{tab:#1}}
\newcommand{\secref}[1]{Section \ref{sec:#1}}

\title{\LARGE \bf
Dense 3D Visual Mapping via Semantic Simplification
}

\author{Luca Morreale and Andrea Romanoni and Matteo Matteucci\\
Politecnico di Milano \\
{\tt\small \{luca.morreale,andrea.romanoni,matteo.matteucci\}@polimi.it}
}

\pgfplotsset{compat=1.14}
\begin{document}

\maketitle
\thispagestyle{empty}
\pagestyle{empty}

\begin{abstract}
Dense 3D visual mapping estimates as many as possible pixel depths, for each image. This results in very dense point clouds that often contain redundant and noisy information, especially for surfaces that are roughly planar, for instance, the ground or the walls in the scene. In this paper we leverage on semantic image segmentation to discriminate which regions of the scene require simplification and which should be kept at high level of details. We propose four different point cloud simplification methods which decimate the perceived point cloud by relying on class-specific local and global statistics still maintaining more points in the proximity of class boundaries to preserve the infra-class edges and discontinuities. 3D dense model is obtained by fusing the point clouds in a 3D Delaunay Triangulation to deal with variable point cloud density.
In the experimental evaluation we have shown that, by leveraging on semantics, it is possible to simplify the model and diminish the noise affecting the point clouds.
\end{abstract}

\section{Introduction}
Dense 3D visual mapping from images aims at building a 3D model which recovers as much part of the scene as possible.
In this field, two main trends emerged: high accuracy dense mapping \cite{vu_et_al_2012,zach2007globally,savinov2016semantic,schoenberger2016mvs,Galliani_2015_ICCV} and dense mapping from sparse data~\cite{labatut2007efficient,litvinov_lhuillier_13,romanoni15b,piazza2018real}.
The former most widespread, methods estimate a model as accurately as possible, computing per-pixel dense correspondences among images and fusing them into the 3D space by  means of patch-based or volumetric representation such as voxels or 3D Delaunay triangulations \cite{zach2007globally,savinov2016semantic,schoenberger2016mvs,Galliani_2015_ICCV}; in some cases a further refinement step is added to improve accuracy and resolution~\cite{vu_et_al_2012,romanoni2017multi}. They are able to produce impressive results, but they are computationally expensive, especially when dealing with large scale environments; they strongly leverage the architecture optimizations, either on GPU or CPU side,  to efficiently compute, move and store data~\cite{cutbuf2016,Zoni:2018:DEO:3212710.3186895}.
On the other hand, methods that are more efficient in terms of space and memory can better estimate a dense map by relying just on sparse data coming from structure from motion methods~\cite{labatut2007efficient,litvinov_lhuillier_13,romanoni15b,piazza2018real}. They are able to run in real-time on a single core of a CPU~\cite{piazza2018real}, without a relevant resources consumption. Nevertheless, they output a low resolution 3D model.

In the former case the resulting model is usually rich of redundancies, especially where the scene presents planar or piecewise-planar regions, while in the latter case 3D points computed by Structure from Motion are too sparse to provide sufficient and satisfying information to reconstruct the scene fine grain details.
Only few works address the problem of trading-off between these two approaches.
Li \etal~\cite{li2016efficient} adapt the resolution of the reconstructed mesh such that regions with less details contain fewer vertices. The method effectively limits the dimension of the mesh, but it is only applied to the mesh refinement stage.
Lafarge \etal~\cite{lafarge2013hybrid} proposed an interesting method which combines the mesh recovered via Multi-view Stereo with high level geometric primitives. This approach simplifies regions of the mesh which fit geometric primitives by means of  a complex jump diffusion scheme acting together with mesh refinement after the initial dense mesh is estimated.
Wu \etal \cite{wu2012schematic} use generalized cylinders and \emph{swept surfaces} to recover a dense simplified structure from Structure from Motion point clouds, while Gallup \etal \cite{gallup20103d} fit a n-layer map into the depth map to produce a compact and robust representation. 

Schindler \etal~\cite{schindler2011classification} fit piece-wise planar surfaces to point clouds to obtain a schematic reconstruction of man-made environments, while
Bourki \etal~\cite{bourki2017patchwork} propose an ad-hoc method for man-made structures: they simplify the depth maps, used to build a patch-based representation, by estimating the vanishing points and by fitting planes on the Structure from Motion output.
Finally, some methods apply planar priors to limit the model noise, but they keep the  model resolution fixed \cite{holzmann2017plane,osman2016patches}.

None of these methods leverages on the semantic information carried by the images which can be effectively estimated by means of semantic image  segmentation methods \cite{visin2016reseg,chen2018deeplab,lin2017refinenet}. Semantics provides very useful priors to reason about which parts of the scene can be simplified and which parts need a more detailed model. Some semantic categories, as the ground or the walls, when reconstructed, contain redundant information which can be simplified, whilst others, as vegetation, require more points to describe their shape with high fidelity.

To the best of our knowledge only the method in Schneider \etal~\cite{schneider2016semantic} proposes a simplified reconstruction that leverage on semantics; the authors jointly optimize the 3D position, the size and the semantic coherence of 3D stixels. However, differently from 3D meshes, stixels are a partial representation of the environment made up by sets of piece-wise planar 3D regions.

\begin{figure}[t]
    \centering
         \includegraphics[width=0.95\columnwidth]{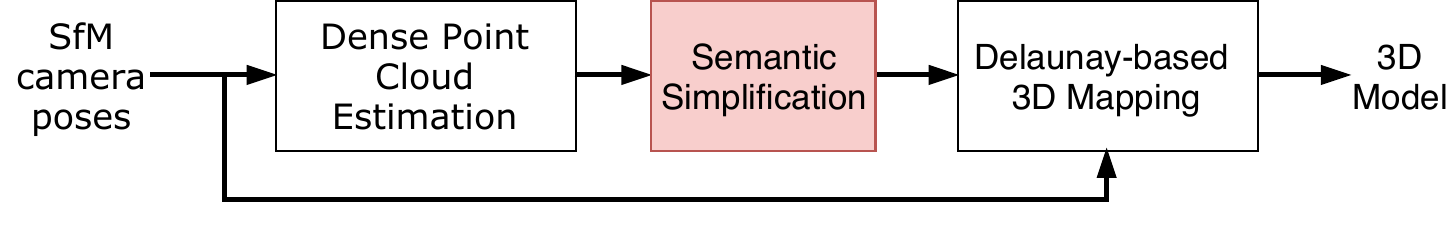}
    \caption{Dense 3D Mapping pipeline.}
    \label{fig:pipeline}
\end{figure}

In this paper we propose a novel method showing how semantics can be effectively be integrated in a 3D mapping pipeline to simplify the dense point clouds (\secref{methods}) before fusing them into a volumetric representation based on 3D Delaunay triangulation (\secref{delaunay}). \figref{pipeline} shows the entire pipeline, we highlighted in red our contribution. We leverage on semantic image segmentation to understand which part of the scene requires less details, \ie, less points, and which shall be maintained as it is. We propose different heuristic to simplify such subsets of the dense point cloud while still accurately representing the scene.
In \secref{experiments} we evaluated our method both quantitatively and qualitatively against two publicly available datasets, proving semantic-based simplification approaches are more effective than non-semantic one. Furthermore, our method is able to significantly reduce the number of points in the reconstruction while keeping its accuracy close, and in some case even better, with respect to the mesh estimated using the point cloud. We also tested the simplification method in an incremental setting suitable for classicals vehicle survey scenarios.

\section{Delaunay Based Mapping}\label{sec:delaunay}
The baseline algorithm we use to build the 3D model of the scene is a batch version of the method in \cite{piazza2018real}, takes as input a set of cameras, a 3D point cloud, and the camera-to-points visibility to estimate a manifold mesh which is consistent with its visibility constrains. Here we briefly describes how the map is estimated, for more information we refer the reader to \cite{romanoni15b} and \cite{piazza2018real}.
Usually this class of methods have been applied to Structure from Motion data \cite{litvinov_lhuillier_13,romanoni15b,piazza2018real}.
In our setting, to obtain a denser and accurate model suitable for optional mesh refinement \cite{vu_et_al_2012}, we densify the input point cloud, combining plane sweeping \cite{hane2014real} and stereo matching \cite{hirschmuller2008stereo} to produce a dense map $DM_i$ for each camera $C_i$ and to estimate the corresponding dense point cloud.

The mesh reconstruction algorithm creates a 3D Delaunay triangulation of the 3D point cloud, and for each tetrahedron $\Delta_i$ it defines $\mu_i^{free}$ and $\mu_i^{matter}$ representing respectively the free-space and matter weights. 
Each visibility ray $Cp$ from camera $C$ to point $p$ is traced  into the triangulation; for each tetrahedron traversed, the value of  $\mu_i^{free}$ is incremented by $1 - e^{\frac{d^2}{2\sigma_{\text{free}}}}$, where $d$ is the distance from $p$ to the closest facet of $\Delta_i$, $\sigma_{\text{free}} = 0.05m$. Then, $Cp$ is extended behind the point $p$ by $10 \sigma_{\text{matter}}m$ and, for each tetrahedron traversed by this extension, the weights $\mu_i^{matter}$ of are incremented by $1 - e^{\frac{d^2}{2\sigma_{\text{matter}}}}$, where $\sigma_{\text{matter}} = 0.01m$

The mesh reconstruction procedure bootstraps from the tetrahedron with the highest free-space vote and it collects iteratively all the nearby free-space tetrahedron such that the manifold property is always fulfilled. One tetrahedron is considered free-space if its votes for free-space are higher than the matter.
The boundary between collected and not collected tetrahedra is the reconstructed 3D manifold model.

\begin{figure}[t]
    \centering
    \includegraphics[width=0.95\columnwidth]{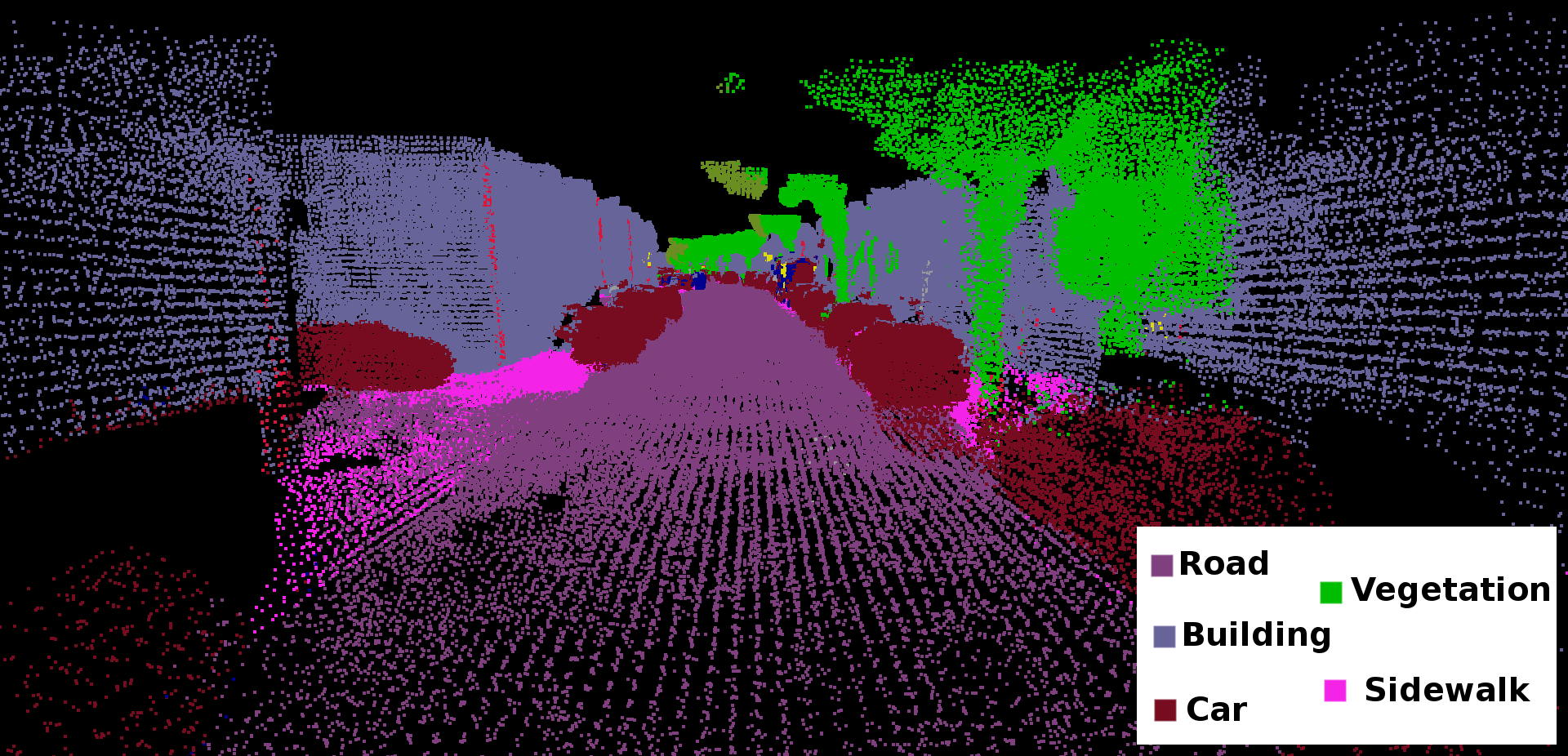}
    \caption{Point Cloud colored accordingly to the semantics.}
    \label{fig:semantic_pc}
\end{figure}


\section{Point Cloud Semantic Simplification} \label{sec:methods}
The output of the dense mapping method from \secref{delaunay} is often redundant especially for some specific areas of the scene; indeed, the ground, which is usually piece-wise planar, ends up to be represented by a relevant number of mesh facets that do not add significant details to the model.

To limit such redundancies and, at the same time, to diminish the number of outliers in the model, we propose to simplify the estimated dense point cloud on which the baseline algorithm builds the 3D Delaunay triangulation instead of post processing the mesh afterward as done in the literature.
To this aim we leverage on semantic image segmentation to discriminate which part of the scene is worth simplifying, in our experiments the ground and the walls, and which part requires a detailed point cloud, and therefore is not affected by the simplification procedure.

To estimate the semantic class of each 3D point $p_i$, we project each point of the dense point cloud, into the segmented image corresponding to the first camera which observed it and we label $p_i$ according to the class of the pixel where the point is projected (see \figref{semantic_pc}).
Although more complex approaches could be employed, \eg, taking the most frequent class among the projection on more images, or weighting the class contribution of each camera depending on the camera-to-point distance, our simple  approach resulted effective in all the experiments and we did not find any improvement with complex ones.

The basic idea behind point cloud semantic simplification is to reduce the number of points belonging to a class composed by piece-wise planar surfaces which likely contain redundant information. 
This decimation procedure can either consider all the points with same semantics at the same time, \ie,  globally, or it can be applied region-wise, \ie, locally. In the former and simpler case, it is only possible to discard points uniformly per class. Instead, if we adopt a more flexible region-wise method we are able to adapt the simplification rate depending on the local point density.

We decided for a local approach and we tested our method with both a fixed radius spherical region and an adaptive region defined by means of the K-Nearest Neighbors (KNN) technique.
The former is more suitable when the size of the scene is approximately known, while the latter in case different regions show widely different densities.
Our approaches are independent on how such areas and neighborhood are defined, therefore, we present here a general region selection algorithm, while in \secref{experiments} we explicitly compare the performance of the two approaches, \ie, spherical search and KNN.

For each point $p_i$ we define a region $\rho_i$ either by K-Neighest Neighbors or radius search. Then we define:
\begin{equation} \label{eq:general_decimation}
    c(\mathcal{D}, \bar{\mathcal{D}}, \tilde{p})  \in [0 ; 1],
\end{equation}
as the \textit{conservation factor}; intuitively it represents the percentage of points in $\rho_i$ maintained after the decimation.
We then define the \emph{region density} $\mathcal{D}$ as the number of points inside $\rho_i$ divided by its area. Since we are considering flat surfaces, when $\rho_i$ is computed by  radius search, the area is defined as the circle having radius equal to the search radius. When $\rho_i$ is computed with a KNN approach the area is equal to the largest rectangle among the faces of the parallelepiped containing its points.
\textit{Average density} $\bar{\mathcal{D}}$  is defined as the average among the density of all the regions belonging to the class of $p_i$, and $\tilde{p}$ is  the percentage of points in the selected area having a different semantics with respect to $p_i$.



\begin{figure}[t]
\centering
  \def\svgwidth{0.98\columnwidth}
  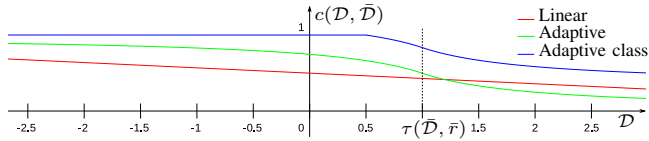
\caption{The conservation factor trend for Linear simplification (with $\bar{c}=0.5$ and $\tau(\bar{\mathcal{D}}, \bar{r})) = 1.0$), Adaptive Simplification ($w=1$) and (with $\tilde{p} = 0.5$ and $\tau(\bar{\mathcal{D}}, \bar{r})) = 1.0$)}
\label{fig:conservation_function}
\end{figure}

We assume that classes requiring simplification, in our case the ground and the walls, represent surfaces which are mostly flat.
For each target class label $l$, we identify a plane over points labeled as $l$ by means of RANSAC and we define its normal as $n_{l}$. Intuitively, we use $n_{l}$ as the target normal for the flat surface we want to reconstruct.
Similarly, we estimate the normal $n_i$ of each point $p_i$ by means of tangent plane approximation \cite{RusuDoctoralDissertation} to formulate the following ranking function:
\begin{equation} \label{eq:ranking}
    r(p_i) = (n_i - {n}_l)^T (n_i - {n}_l) .
\end{equation}
The closer $n_i$ is to ${n}_l$, the smaller $r(p_i)$.

We use \textit{ranking score} $r(p_i)$ to choose which subset of points is kept, \ie, for each region $\rho_i$ we rank all the points $p$ according to the \textit{ranking score} $r(p_i)$ and we discard the last $n = (1-c) \cdot |\rho_i|$, where $|\rho_i|$ represents the number of points in $\rho_i$.
The outcome of the point simplification depends on how we define the conservation factor $c$. We here propose four different approaches that we discuss and evaluate in the experimental section (Section \ref{sec:experiments}).

\subsubsection*{Linear Simplification (LS)}
The simplest simplification method is based on a linear conservation factor; it is based on the region and average densities as follows:
\begin{equation} \label{eq:linear_decimation}
    c(\mathcal{D}, \bar{\mathcal{D}}) = -\frac{\bar{\mathcal{D}}}{\bar{c}} * (\mathcal{D} - \bar{\mathcal{D}}) + \bar{c} ,
\end{equation}
where $\bar{c}$ is the target conservation factor for an area with average density (\eg, red line in \figref{conservation_function}). 
Lets consider for instance $\bar{c}=0.4$; the idea of LS is to keep around $40\%$ of the points in the areas where the density is similar to the average density and to retain more than $40\%$ of points for less dense areas. 

\begin{figure}[t]
    \centering
     \setlength{\tabcolsep}{5px}
    \begin{tabular}{ccc}
         \includegraphics[width=0.3\columnwidth]{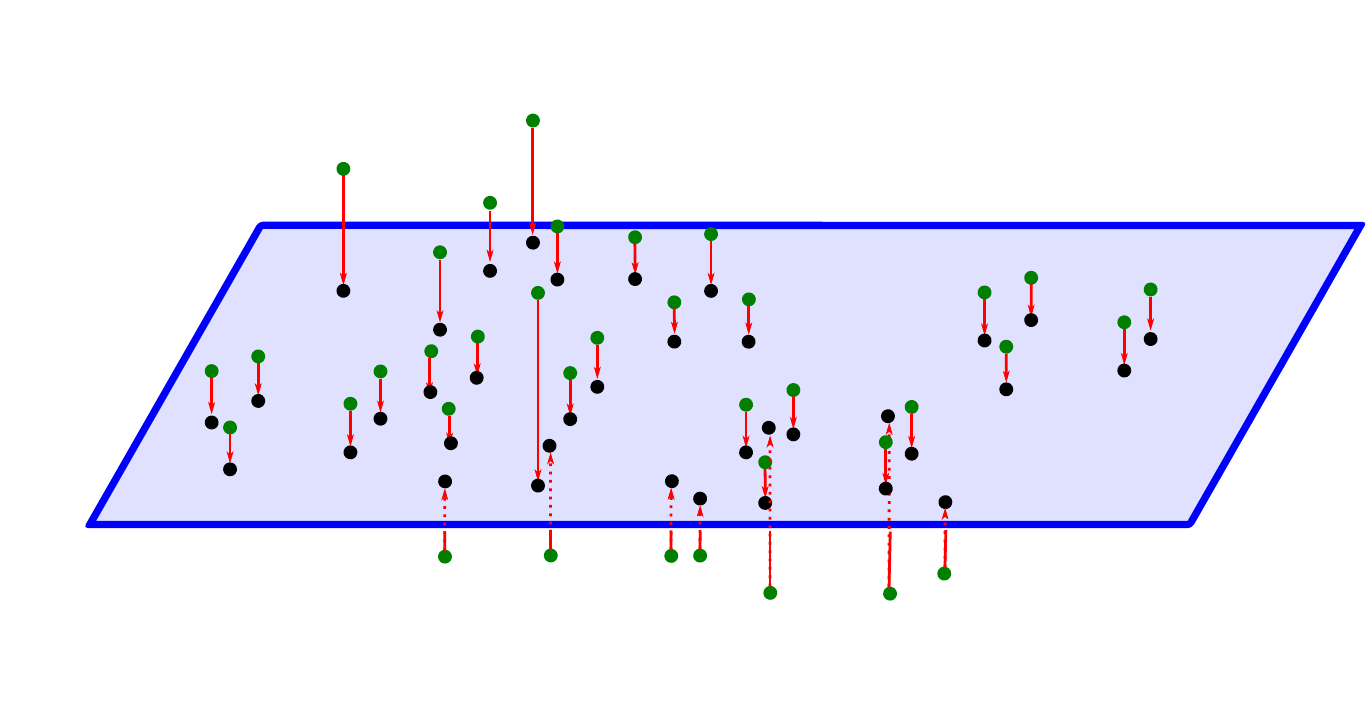} &
         \includegraphics[width=0.3\columnwidth]{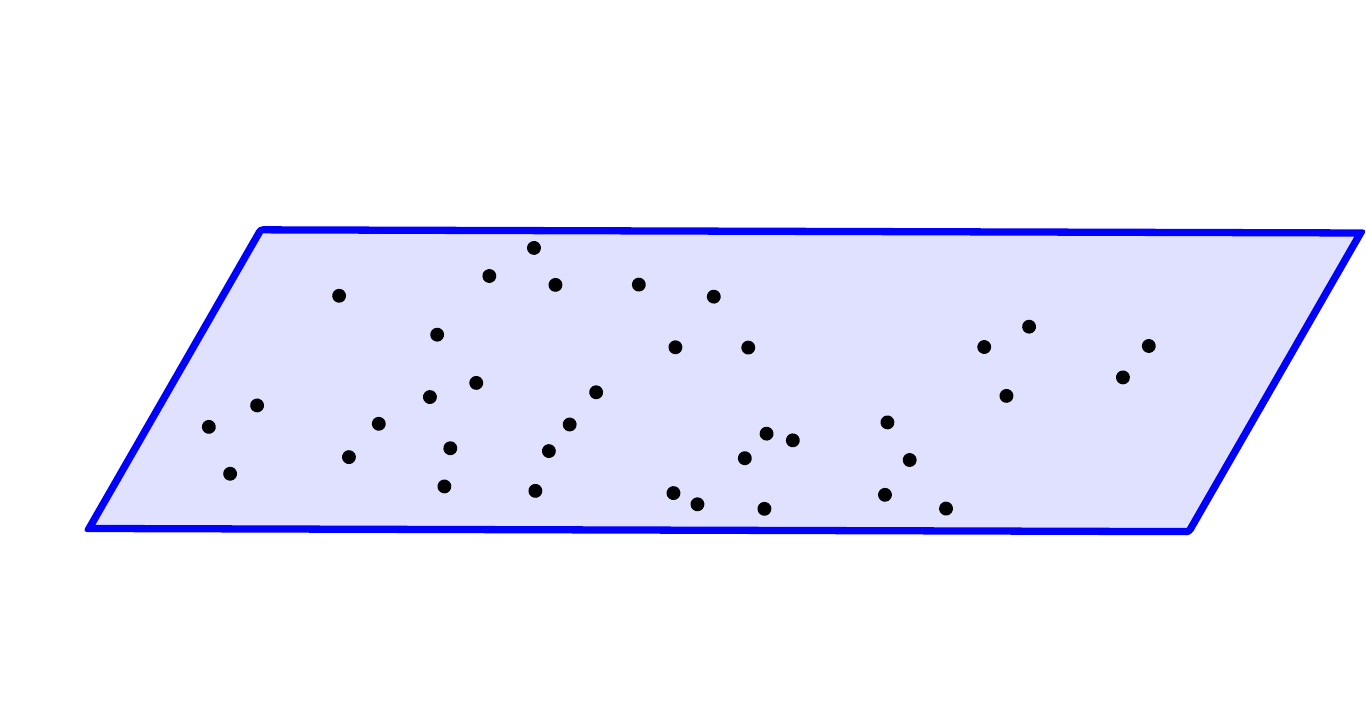} &
         \includegraphics[width=0.3\columnwidth]{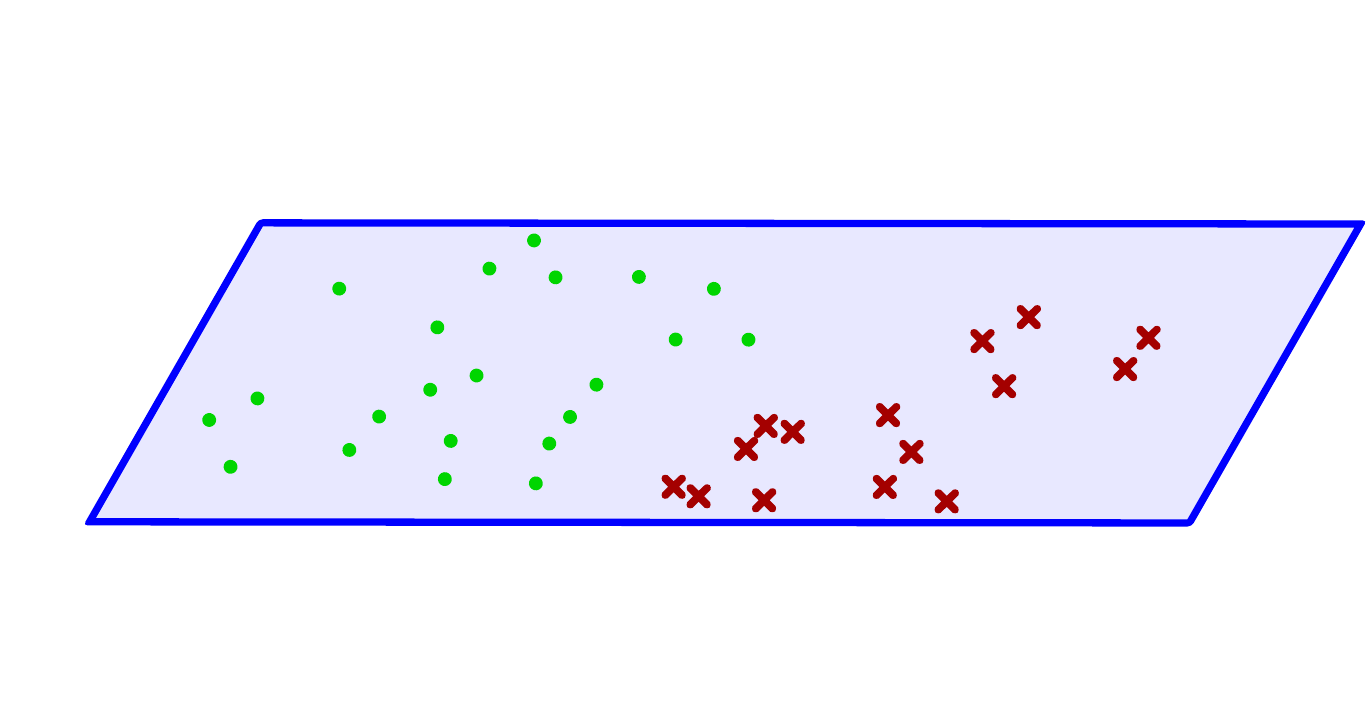} \\
         \textbf{(a)} & \textbf{(b)} & \textbf{(c)} \\
         \includegraphics[width=0.3\columnwidth]{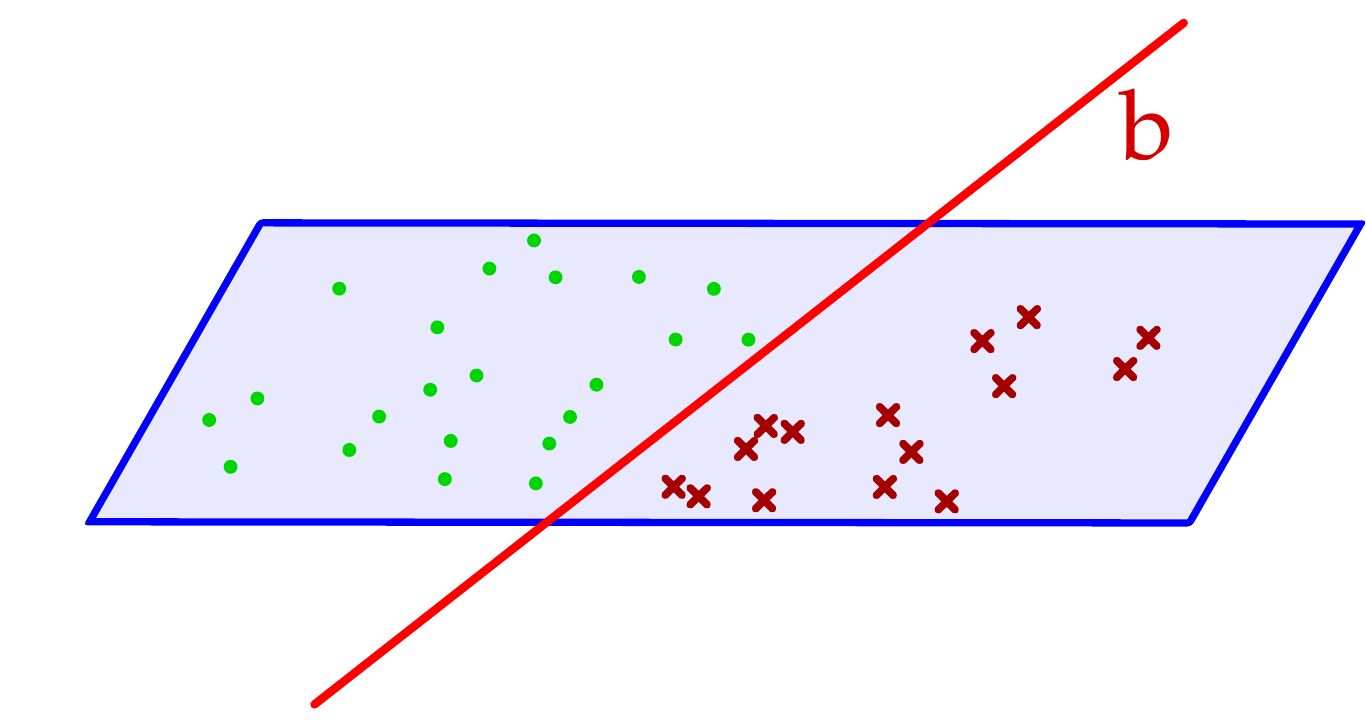} &
         \includegraphics[width=0.3\columnwidth]{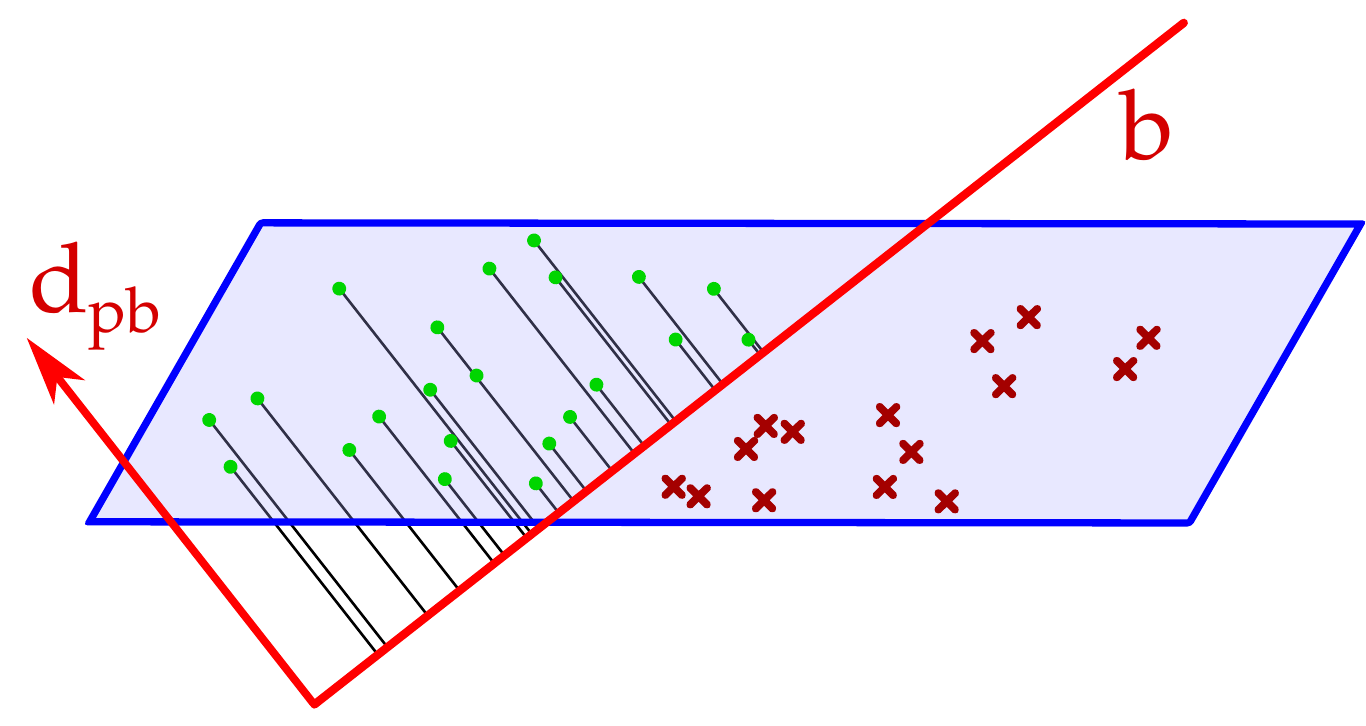} &
         \includegraphics[width=0.3\columnwidth]{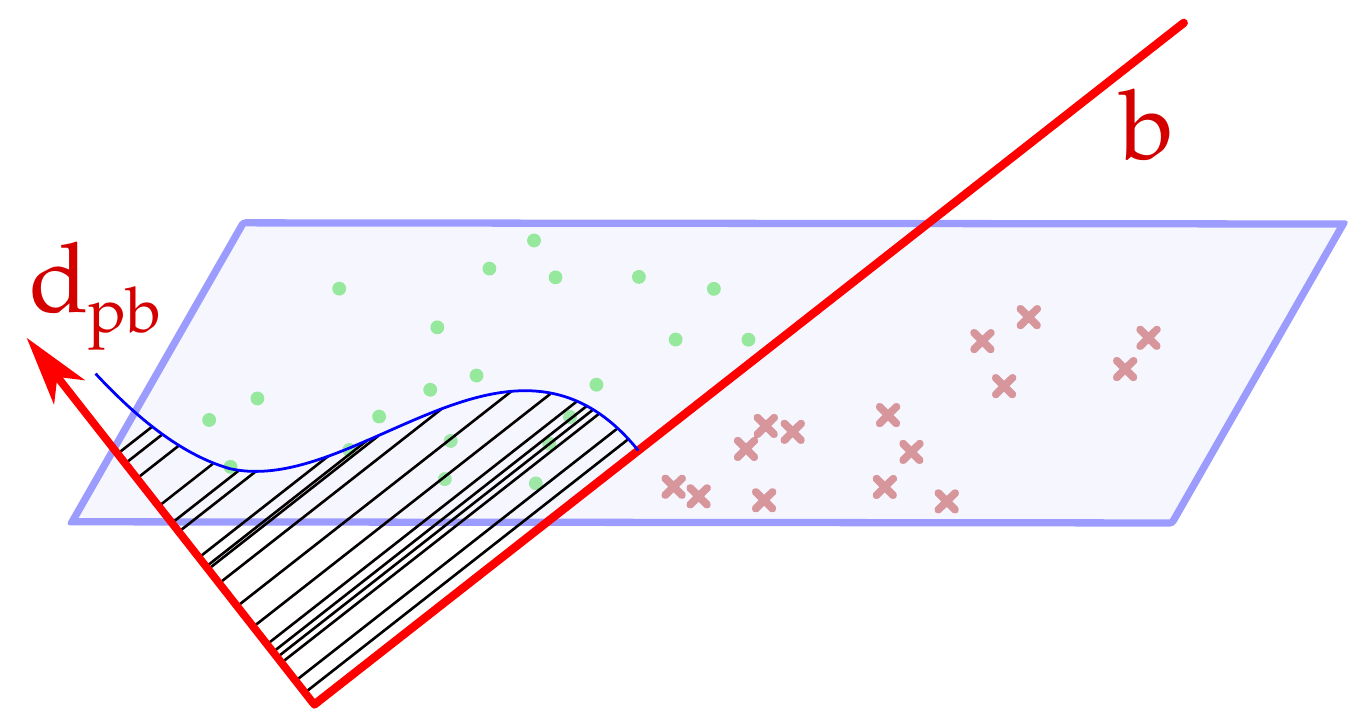} \\
         \textbf{(d)} & \textbf{(e)} & \textbf{(f)} \\
    \end{tabular}
    \caption{
     Step by step procedure to derive $P_B$ for a point P.  We project the points on the plane (a-b), we assign a binary label that indicates if a point belongs to the class of the point that generated the region   (c) we estimate the separating line (d) we project them to the axis $d$ (e) to compute the point probability of being maintained or discarded (e).  
    }
    \label{fig:projection_steps}
\end{figure}

\begin{figure*}[t]
    \centering
    \setlength{\tabcolsep}{1px}
    \begin{tabular}{ccc}
         \multicolumn{3}{c}{\includegraphics[width=0.9\columnwidth]{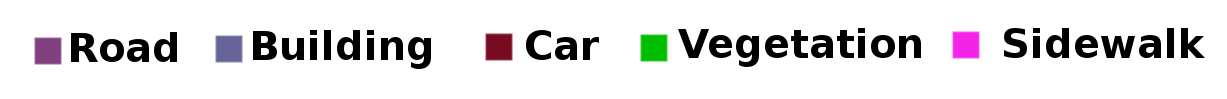}} \\
         \includegraphics[width=0.64\columnwidth,height=0.35\columnwidth]{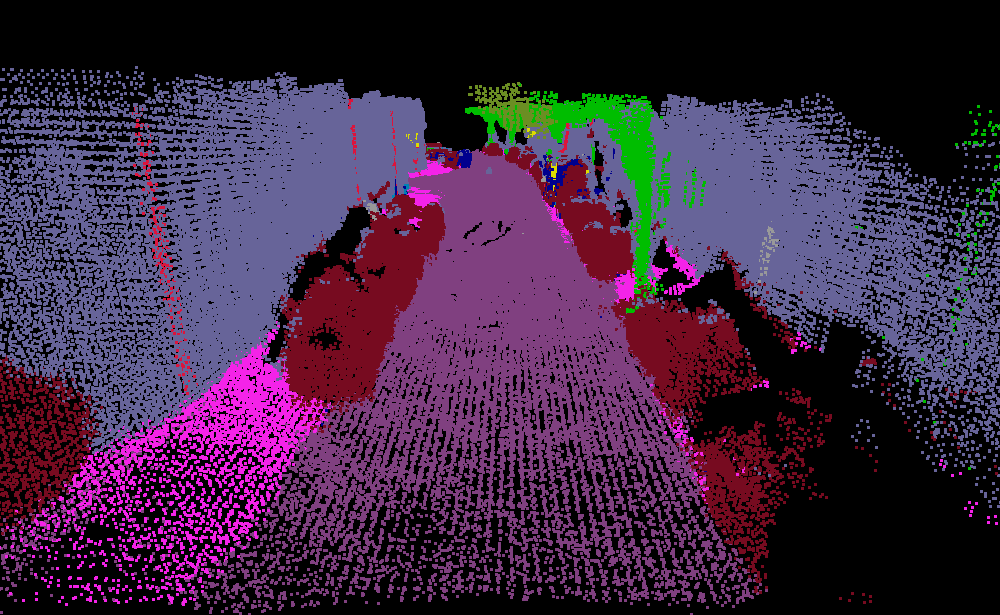} &
         \includegraphics[width=0.64\columnwidth,height=0.35\columnwidth]{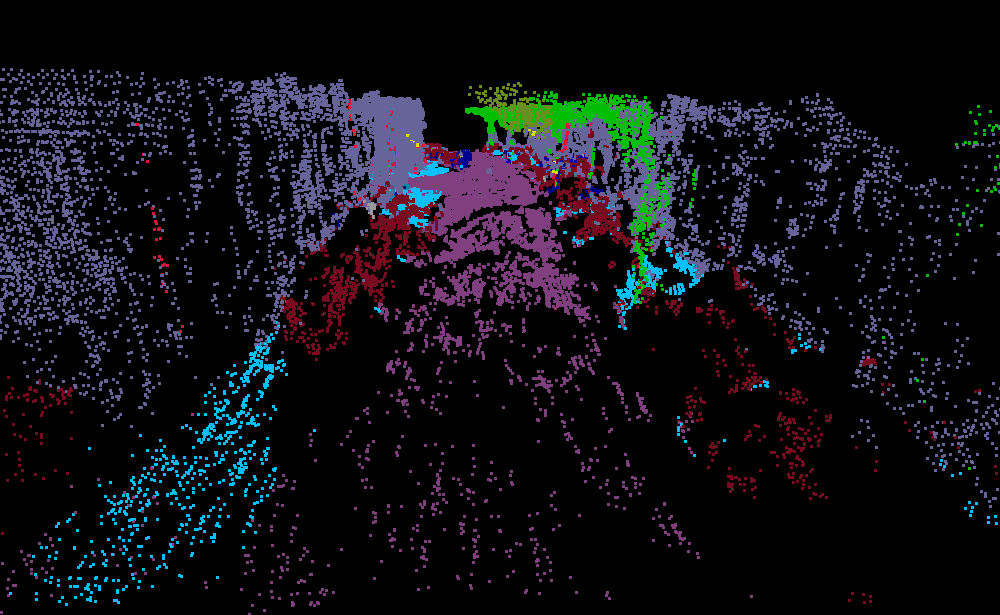} &
         \includegraphics[width=0.64\columnwidth,height=0.35\columnwidth]{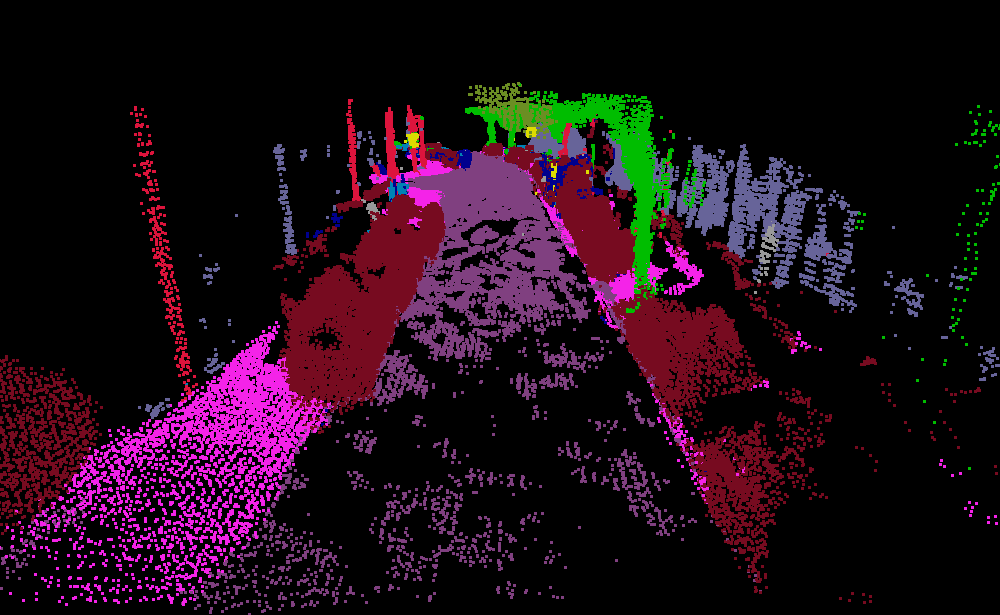} \\
         \textbf{(a)} Original  & \textbf{(b)} Baseline Simplification (BPS) & \textbf{(c)} Linear Simplification\\
         
         \includegraphics[width=0.64\columnwidth,height=0.35\columnwidth]{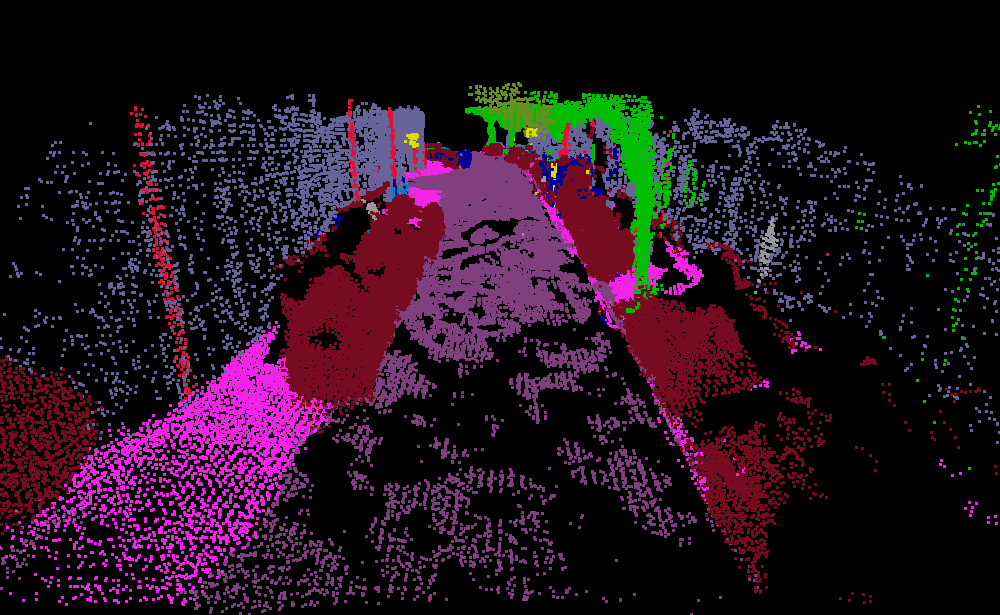} &
         \includegraphics[width=0.64\columnwidth,height=0.35\columnwidth]{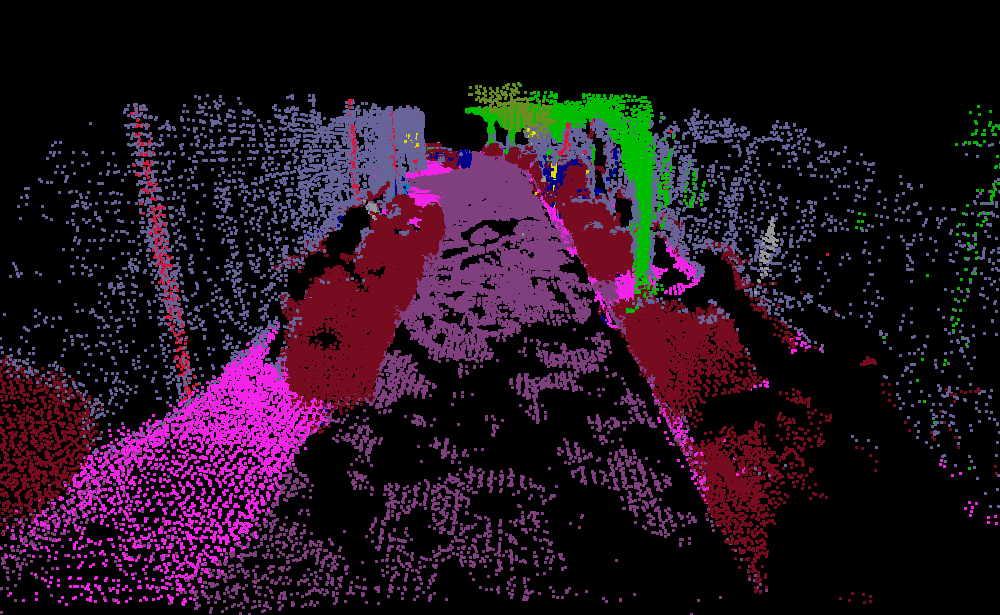} &
         \includegraphics[width=0.64\columnwidth,height=0.35\columnwidth]{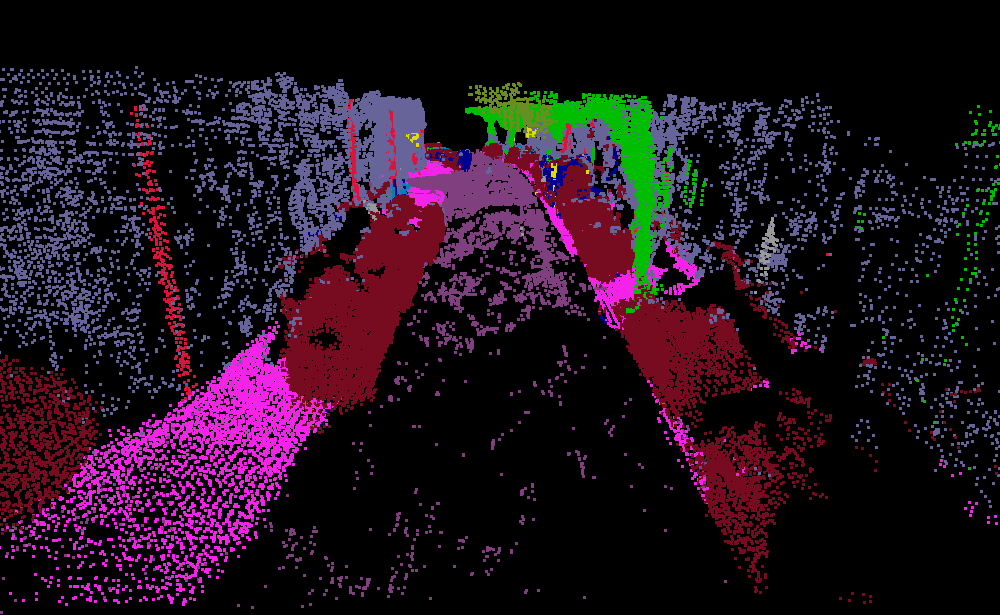} \\
         \textbf{(d)} Adaptive Simplification  & \textbf{(e)} Adaptive Class Simplification & \textbf{(f)} Probabilistic Simplification\\
         

    \end{tabular}
    \caption{Street perspective of the KITTI point clouds.}
    \label{fig:simplified_clouds_view}
\end{figure*}



\subsubsection*{Adaptive Simplification (AS)}
To have a higher simplification where the point density diverges with respect to the average of the point cloud density, in Adaptive Simplification we use a function $f$ similar to the sigmoid function (green curve in \figref{conservation_function}):
\begin{equation} \label{eq:atan_like}
    f(x) = 0.5 + \frac{w  x}{2  (1 + w  |x|)} ,
\end{equation}
where $w$ is a stretching factor: a small $w$ corresponds to larger  difference between two similar densities.
We formalize the conservation factor with:
\begin{equation} \label{eq:adaptive_decimation}
    c(\mathcal{D}, \bar{\mathcal{D}}) = f(\mathcal{D} - \tau(\bar{\mathcal{D}}, \bar{r})) .
\end{equation}
where $\tau(\bar{\mathcal{D}}, \bar{r})$ is defined such that $f(\bar{\mathcal{D}}) = \bar{c}$. 


\subsubsection*{Adaptive Class Simplification (ACS)}
When the region $\rho_i$ contains points belonging to different classes, we might prefer to limit the simplification since it is more likely that the region is close to the boundary of two or more semantic regions. The points near to the class boundaries may correspond to changes in surface orientation; therefore, they are useful to define the right shape of the reconstructed model.
To do so, with \textit{Adaptive Class Simplification}, we remove the points depending on the semantically different points contained in $\rho_i$. 
In this case we define $c$ by means of  $\mathcal{D}$, $\bar{\mathcal{D}}$ and $\tilde{p}$, \ie, the percentage of the points belonging to other classes with respect to the one $p_i$ belongs to.
Equation \eqref{eq:adaptive_decimation} becomes:
\begin{equation} \label{eq:class_adaptive_decimation}
    c(\mathcal{D}, \bar{\mathcal{D}}, \tilde{p}) = (1 + \tilde{p}) f(\mathcal{D} - \tau(\bar{\mathcal{D}}, \bar{r})) ,
\end{equation}
when $f$ is close to $1$ and $\tilde{p}$ is not zero, $c$ might exceed $1$, to avoid this issue we saturate it imposing $c = \min(1, c(\mathcal{D}, \bar{\mathcal{D}}, \tilde{p}) )$.
Intuitively, when $\tilde{p}$ is small the area contains few points of other classes, hence it slightly increases $c$ and the behaviour is similar to the adaptive simplification. On the other hand, when $\tilde{p}$ is close or above $0.5$, \ie, the center of the region is close to the boundary, $c$ is largely increased, and consequently we remove significantly less points. 
For instance in \figref{conservation_function} we plot the trend of Equation \eqref{eq:class_adaptive_decimation} when $\tilde{p} = 0.5$.


\subsubsection*{Probabilistic simplification (PS)}
Differently from previous approaches we define the probability $P_c$ that a point $p_i$ is conserved. Moreover,  while in previous methods we extracted a region for each point, in this case, to limit the computational burden, we compute a predefined number of regions $N_r$ experimentally fixed to $\frac{|\rho|}{8}$, where $|\rho|$ is the size of the cloud. The regions are defined around $N_r$ points randomly chosen among those belonging to the class $l$ we are going to simplify.
Even if in principle this choice may leave some area uncovered by the simplification method, experiments showed that in practice this issue is not statistically relevant.

We define $P_c$ as the composition of two probabilities:
\begin{multline} \label{eq:probabilistic_conservation}
    P_c(p | \tilde{p}, \mathcal{D}, \bar{\mathcal{D}}, \rho_i) = \\
    i(\tilde{p}) \cdot P_I(p | \mathcal{D}, \bar{\mathcal{D}}) + (1 - i(\tilde{p})) \cdot P_B(p | \mathcal{D}, \bar{\mathcal{D}}, \rho_i) ,
\end{multline}
where $i(\tilde{p})$ is an indicator function defined as follows:
\begin{equation}
    i(\tilde{p}) =
    \begin{cases}
        0 & \tilde{p} \geq 0.1 \\
        1 & \text{otherwise} \\
    \end{cases} .
\end{equation}
Intuitively, when the percentage of points with a different semantics is less then $10\%$, \ie, when the region  $\rho_i$ is far from the inter-class boundary, only the probability $P_I$ is active, otherwise, \ie, when $\rho_i$ is close to the boundary, $P_c(p | \tilde{p}, \mathcal{D}, \bar{\mathcal{D}}, \rho_i) = P_B$.

The probability $P_I$ represents the idea that the closer a point is to the center of a flat surface, the lesser it is useful.
To formalize $P_I$ we project every point $p\in\rho_i$ to the plane $\pi_l$, having normal $n_{l}$. For every point we then used a gaussian distribution, $\mathcal{N}$, to estimate the importance of the point in terms of probability.  
Specifically, we define:
\begin{equation}
    P_I(p | \mathcal{D}, \bar{\mathcal{D}}) = 1 - \mathcal{N}(\mu, \Sigma) ,
\end{equation}
where $\mathcal{N}$ is the normal probability distribution depending on $\mu$, \ie, the projection of $p_i$ on $\pi_l$ and $\Sigma$, \ie, a diagonal matrix formalized as:
\begin{equation}
    \Sigma = 
    \begin{bmatrix}
         \tilde{d} \cdot \sigma^2 &              0          \\
                     0          &  \tilde{d} \cdot \sigma^2 \\
    \end{bmatrix} ,
\end{equation}
where $\tilde{d}$ is the ratio of $\mathcal{D}$ and $\bar{\mathcal{D}}$, and $\sigma$ is an a-priori defined standard deviation.

When a region is close to the class boundaries it may happen that a point in the center of the region is near to the boundary. In this case we want to keep such points. For this reason we need to explicitly detect the boundary and then remove points far from it.
To this extent we formulate $P_B$.
We consider all the points inside the region $\rho_i$ and we project them on the plane $\pi_l$ fitted to the points (\figref{projection_steps}(a-b)). Let $l$ be the class considered; locally the points belonging and not belonging to $l$  are usually separated by a line $\mathbf{b}$ which is roughly the class boundary (\figref{projection_steps}(c)). We find $\mathbf{b}$ as a linear classification problem through a Support Vector Machine (SVM) with linear kernel (\figref{projection_steps}(d)).
For each point $p$ we compute the distance $d_{p\mathbf{b}}$ from the estimated boundary $\mathbf{b}$ and we compute the probability of point conservation $P_B(p)$ (\figref{projection_steps}(f)) as:
\begin{equation}
    P_B(p) = P_B(d_{p\mathbf{b}}) \sim \mathcal{N}(0, \sigma^2),
\end{equation}
the closer a point to the boundary the more important it is. Therefore, to assign the maximum importance to points close to the boundary we center the Gaussian distribution on the line, \ie, with distance 0. The point importance decreases as the probability. 
Once  $P_c$ is known, we remove a point from the cloud if, extracting a random number, it is greater than $P_c(p | \tilde{p}, \mathcal{D}, \bar{\mathcal{D}}, \rho_i)$.



\begin{figure}[t]
    \centering
    \pgfplotsset{scaled y ticks=false}
    \begin{tabular}{c}
\begin{tikzpicture}[scale=0.65]
\begin{axis}[%
    scatter/classes={%
        baseline={mark=*,draw=darkorchid, fill=darkorchid},
        proposed={mark=o,draw=brightgreen, fill=brightgreen},
        original={mark=triangle,draw=blue, fill=blue}
    },
    mark=*, 
    only marks, 
    scatter src=explicit symbolic,
    nodes near coords*={\Label},
    visualization depends on={value \thisrow{label} \as \Label}, 
    xlabel = {Mesh size (in MB)},
    ylabel = {RMSE (in m)}
]
\draw [solid, color=cadmiumgreen] (13.7,0) -- (13.7,1);

\addplot[scatter,only marks,%
    scatter src=explicit symbolic]%
table[meta=class] {
x y class label
34.1 0.7843 original O
16.9 0.7946 baseline G(B)
25.0 0.7846 baseline LS(B)
15.0 0.7830 baseline AS(B)
11.2 0.8093 baseline PS(B)
18.5 0.7740 proposed LSR
16.7 0.7707 proposed LSK
20.0 0.7819 proposed ASR
13.9 0.7939 proposed ASK
20.8 0.7879 proposed ACSR
16.1 0.8102 proposed ACSK
19.2 0.7797 proposed PS
    };

\end{axis}
\end{tikzpicture}
\\
(a) KITTI\\
\begin{tikzpicture}[scale=0.65]
\begin{axis}[%
    scatter/classes={%
        baseline={mark=*,draw=darkorchid, fill=darkorchid},
        proposed={mark=o,draw=brightgreen, fill=brightgreen},
        original={mark=triangle,draw=blue, fill=blue}
    },
    scatter, mark=*, 
    only marks, 
    scatter src=explicit symbolic,
    xlabel = {Mesh size (in MB)},
    ylabel = {RMSE (in m)}
]
\draw [solid, color=cadmiumgreen] (13.5,0) -- (13.5,1);
\addplot[scatter,only marks,%
    scatter src=explicit symbolic,
 ]%
table[meta=class] {
x y class
34.5 0.0524 original
16.0 0.0518 baseline
39.4 0.0538 baseline
34.2 0.0534 baseline
34.0 0.0522 baseline
17.8 0.0694 proposed
19.2 0.0165 proposed
16.7 0.0666 proposed
18.3 0.0629 proposed
19.2 0.0671 proposed
21.2 0.0614 proposed
16.0 0.0673 proposed
};
\node[label={90:{O}},inner sep=2pt] at (axis cs:34.5, 0.0524) {};
\node[label={90:{G(B)}},inner sep=2pt] at (axis cs:16.0, 0.0518) {};
\node[label={90:{LS(B)}},inner sep=2pt] at (axis cs:39.4, 0.0538) {};
\node[label={180:{AS(B)}},inner sep=2pt] at (axis cs:34.2, 0.0534) {};
\node[label={270:{PS(B)}},inner sep=2pt] at (axis cs:34.0, 0.0522) {};
\node[label={90:{LSR}},inner sep=2pt] at (axis cs:17.8, 0.0694) {};
\node[label={90:{LSK}},inner sep=2pt] at (axis cs:19.2, 0.0165) {};
\node[label={270:{ASR}},inner sep=2pt] at (axis cs:15.8, 0.0666) {}; 
\node[label={270:{ASK}},inner sep=2pt] at (axis cs:18.3, 0.0629) {};
\node[label={0:{ACSR}},inner sep=2pt] at (axis cs:19.2, 0.0671) {};
\node[label={0:{ACSK}},inner sep=2pt] at (axis cs:21.2, 0.0614) {};
\node[label={180:{PS}},inner sep=2pt] at (axis cs:16.0, 0.0673) {};

\end{axis}
\end{tikzpicture}

\\
(b) fountain-P11
    \end{tabular}
    \caption{Mesh statistics for each method: in blue the original model (O), in purple the baselines and in green the proposed methods. }  
    \label{fig:results_plot}
\end{figure}
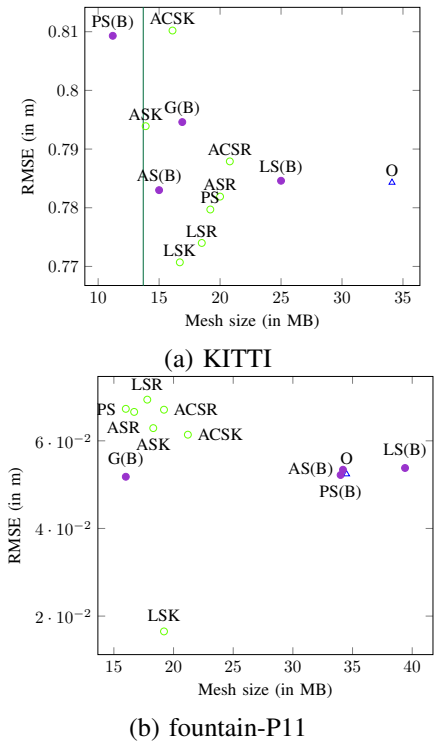


\section{Experiments} \label{sec:experiments}

To prove the effectiveness of the proposed semantic-based simplification methods, and demonstrate semantic approaches are superior to non-semantic one, we test them against several baselines which do not rely on semantics. 
In the first group of baselines we consider the whole point cloud as if of single class and we simplify with the methods proposed in \secref{methods}, \ie, without semantics. The Linear and Adaptive Simplifications act on the point cloud purely based on the region density, while in the Probabilistic method the $P_B$ term is neglected since there are no more boundaries. 
A second baseline is the purely geometric approach proposed in~\cite{cignoni1998comparison}, which, by relying on surface curvature, focuses the simplification on flat surfaces.

\subsection{Simplification Results}
\begin{figure}[t]
    \centering
     \setlength{\tabcolsep}{2px}
    \begin{tabular}{cc}
         \includegraphics[width=0.484\columnwidth]{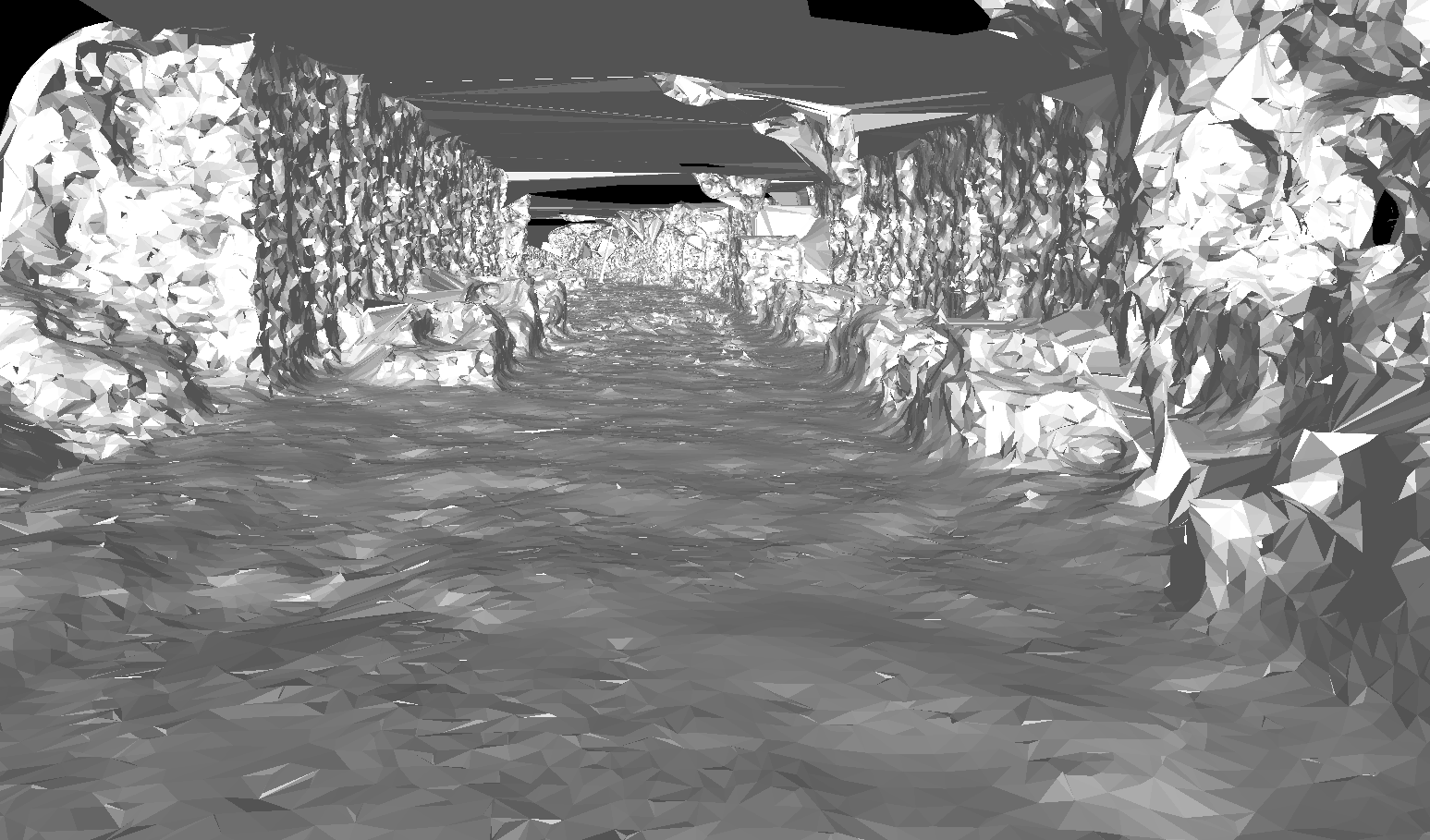} &
         \includegraphics[width=0.484\columnwidth]{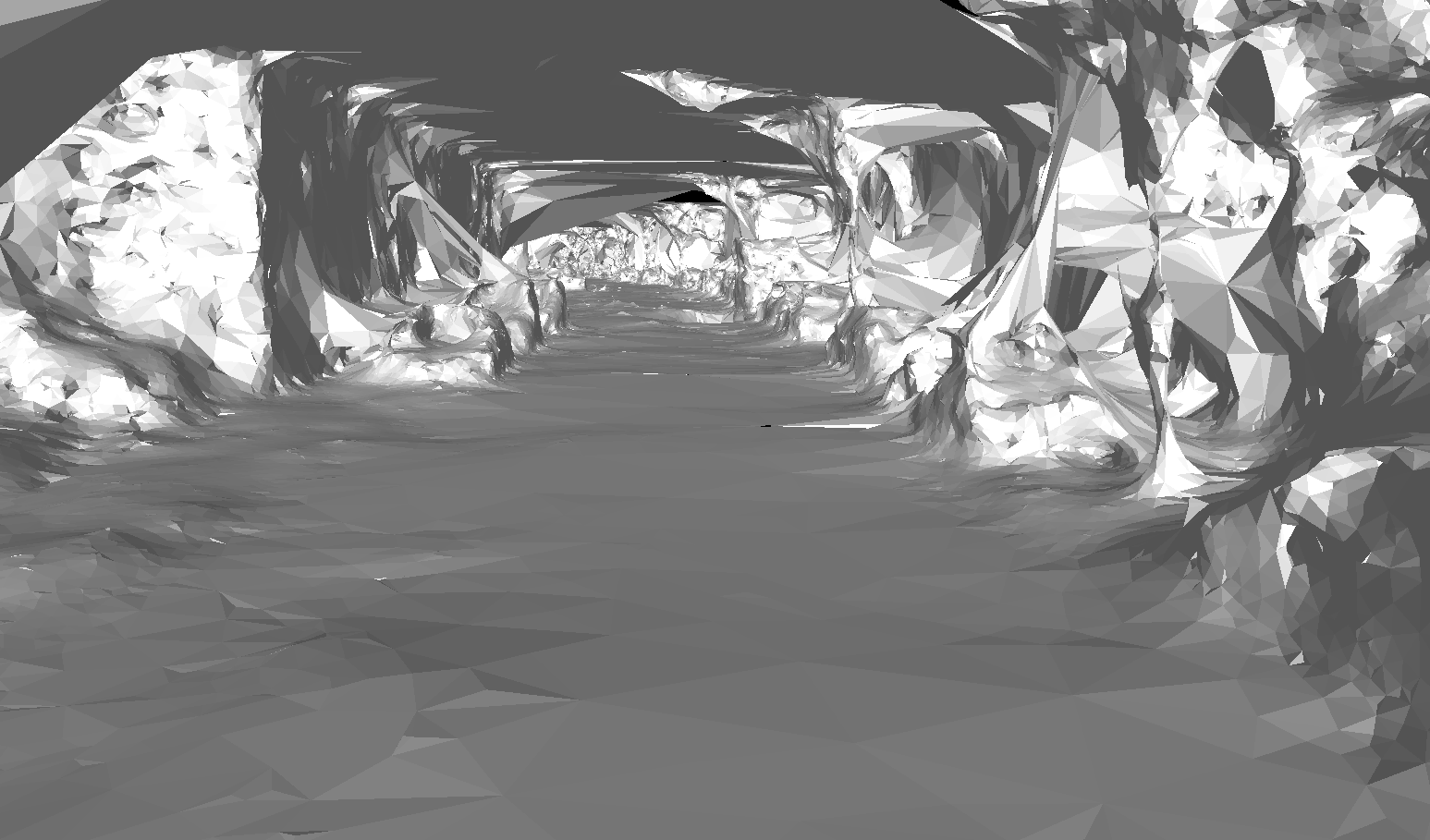} \\
         \textbf{(a)} Original &
         \textbf{(b)} Simplified
    \end{tabular}
    \caption{Reconstruction of the KITTI dataset. (a) shows the original reconstruction, while (b) the simplified version using the Probabilistic simplification. Let notice, how more smooth the simplified version is.}
    \label{fig:meshes_view}
\end{figure}

We evaluated our simplification methods against the sequence 95 of the KITTI dataset \cite{geiger_et_al12}  and the fountain-P11 of EPFL dataset \cite{strecha2008}.
In the former case, stereo pairs are provided, therefore we generate a dense point cloud by means of semi-global block matching \cite{hirschmuller2008stereo}; in the latter case, we adopt the plane sweeping algorithm ~\cite{hane2014real} to the 11 high-resolution images provided. Then, for each image we choose to predict the semantic segmentation by means of ReSeg \cite{visin2016reseg} for the KITTI dataset and MultiBoost~\cite{webb2000multiboosting} for the fountain dataset (since a very limited set of images is available). In principle whatever image segmentation method is applicable.
In the experiments we simplify these two semantic point clouds and we evaluate how this impacts the accuracy of the recovered surfaces. In particular, for the KITTI dataset, we evaluate the accuracy of the reconstructed model, for each camera, by comparing the  points of the Velodyne projected into the image plane with the corresponding depth map generated by the 3D model. For the fountain-P11 dataset we follow the evaluation procedure presented in the paper \cite{strecha2008} and we compare the depth map generated by the dense ground truth point cloud and by the 3D model. The accuracy measure we use is the Root Mean Squared Error (RMSE) of the depth values (where depths are valid).
\figref{results_plot}(a) shows the statistics about the reconstructed meshes in the KITTI dataset in terms of disk occupancy (the mesh is stored as an OFF file without compression) with respect to the reconstruction accuracy. The blue mark represents the original mesh (O) the green marks represent the proposed methods, \ie, LS, AS, ACS, PS, where the final letter K or R, denotes the search method used, \ie, KNN or Radius. Finally, the purple points are the baselines: Baseline LS, AS, PS and the baseline based on geometry curvature (BG).
Both baselines and proposed approaches significantly reduce the size of the reconstruction with respect to the original. In particular, the number of points belonging to ground and walls, \ie, the two classes involved in the simplification procedure, decreases up to two orders of magnitude.
However, while the accuracy of the baselines is in general slightly worse than the original, the proposed linear and probabilistic methods consistently improve the accuracy of the reconstruction. Indeed, they  provide a good trade-off between mesh over-simplification and the removal of redundant and noisy data (see \figref{meshes_view}).

In \figref{results_plot}(a) the green vertical line represents the size achievable by removing all the ground and wall points. Therefore, a class-aware method too close to the line is over-simplifying, whilst a method too distant from it is under-simplifying.
Since the baselines simplifies the whole mesh and therefore are not limited to ground and wall, they can cross this line.


As a qualitative comparison, we refer the reader to the supplementary video and we illustrate in \figref{simplified_clouds_view} the original point cloud colored according to semantics, compared with the Baseline Probabilistic Simplification (BPS) and the other proposed simplification approaches. 
As expected, BPS decimates the point cloud evenly over the entire scene while PS, \figref{simplified_clouds_view}(f), is able to keep the details of the trees and the cars and it significantly simplifies the walls and the ground. For this reason the 3D mapping from Probabilistic Simplification point cloud is sensibly better than the one obtained via BPS which suffers from over-smoothing.

\begin{figure*}[t]
    \centering
    \setlength{\tabcolsep}{2px}
    \begin{tabular}{ccc}
         \multicolumn{3}{c}{\includegraphics[width=0.6\columnwidth]{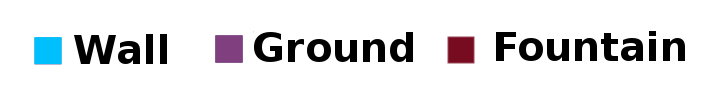}} \\
         \includegraphics[width=0.6\columnwidth,height=0.35\columnwidth]{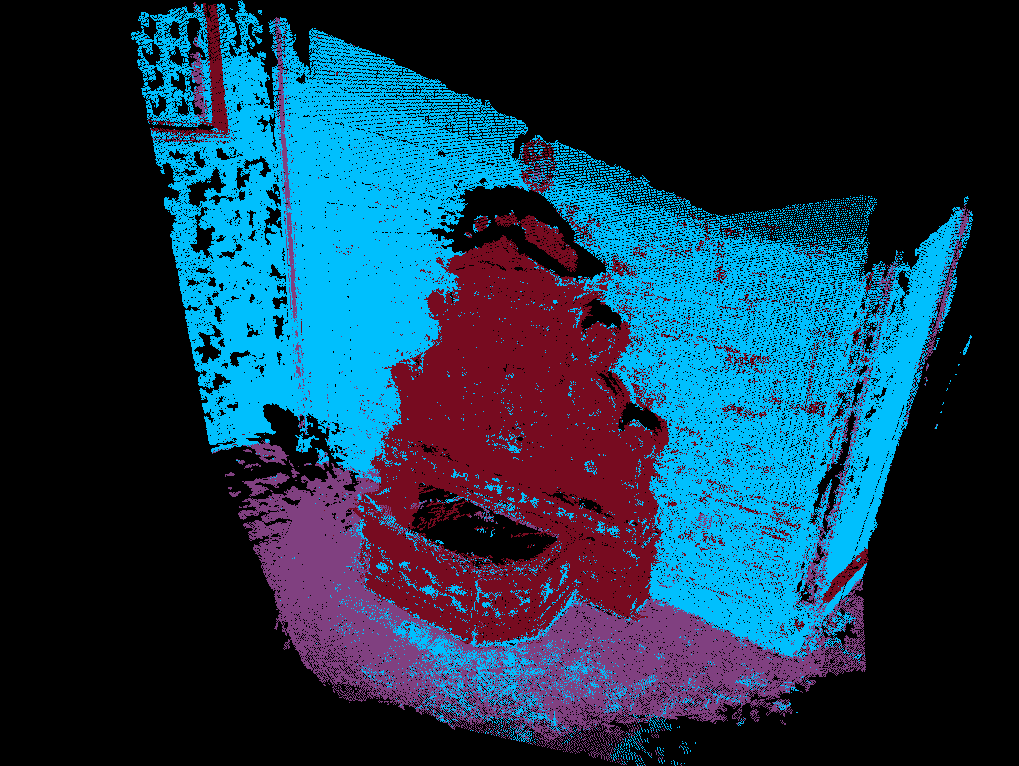} &
         \includegraphics[width=0.6\columnwidth,height=0.35\columnwidth]{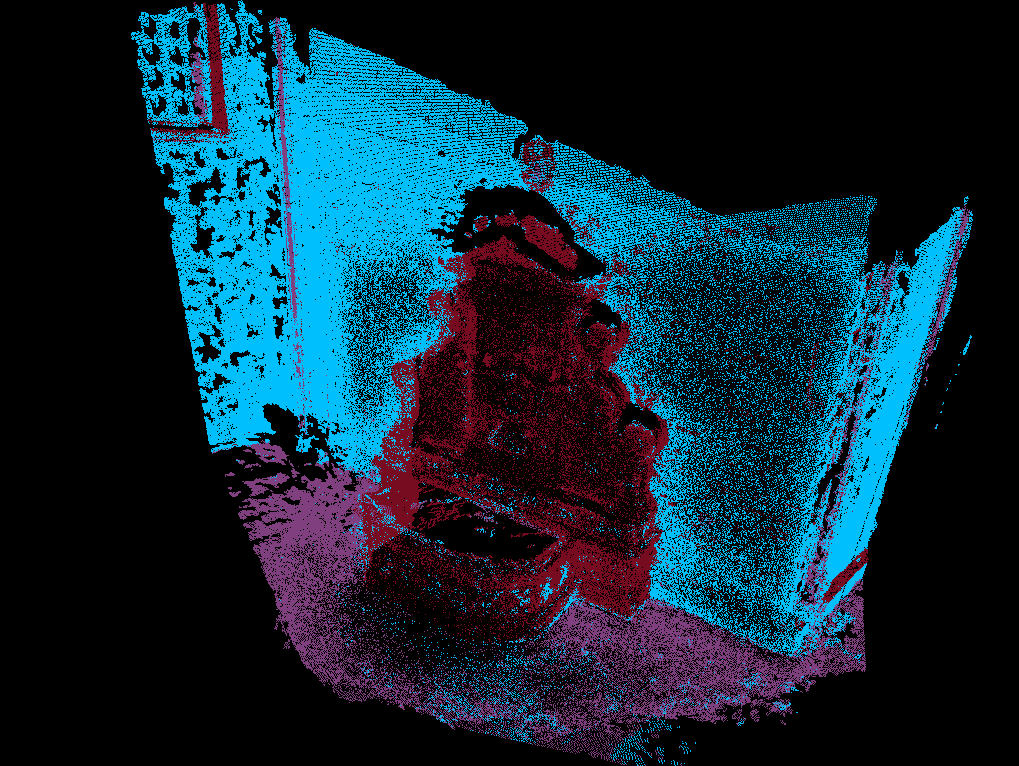} &
         \includegraphics[width=0.6\columnwidth,height=0.35\columnwidth]{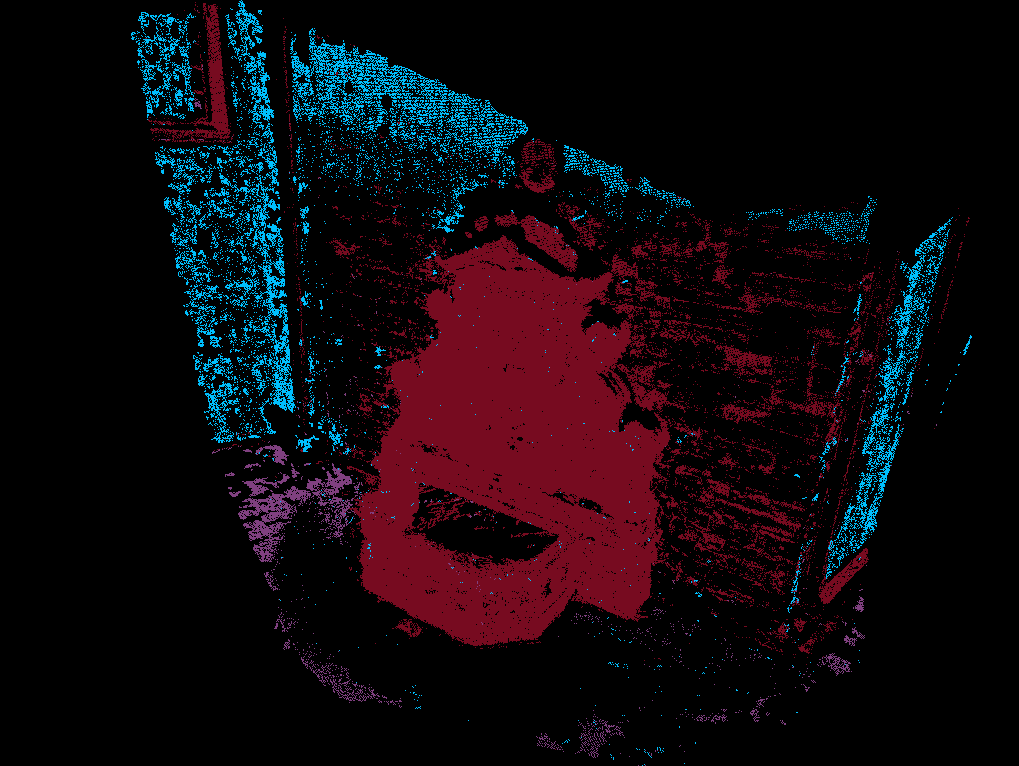} \\
         \textbf{(a)} Original  & \textbf{(b)} Baseline Simplification (BPS) & \textbf{(c)} Probabilistic Simplification \\
    \end{tabular}
    \caption{Frontal perspective of the fountain-P11 point clouds. }
    \label{fig:simplified_clouds_view_fountain}
\end{figure*}

We tested our approaches also with the fountain-P11 dataset and \figref{results_plot}(b) shows the reconstruction statistics: the Linear simplification method achieves the best result, while the other proposed approaches achieve similar performances among each other and they are close to the Baseline Geometric simplification. Other baseline methods are overcome.
In this case the dataset is significantly different with respect to KITTI, especially because the fountain contains many details and its density is much higher than the ground and the wall. Then, even if a simplification procedure is applied to that region, the mapping accuracy does not drop significantly.

Therefore, baselines can simplify the highly redundant surface of the fountain, without virtually loss in accuracy, while the proposed method simplifies only walls and ground. To have a fairer comparison, we tested PS combined with a random simplification over the fountain. We randomly retained $40\%$ of the points composing the fountain. The proportion has been chosen accordingly to the percentage of points retained by the BG. As final model we obtained a smaller 3D model, $13 MB$ while both PS and BG originally achieved around $16 MB$, furthermore modified PS accuracy reaches $0.0655m$ RMSE error, which is similar to BG and PS.

Regarding the region selection method, radius search, proved to be easier to tune and to use in both scenarios with respect to KNN, indeed a good initial radius search can be estimated knowing the size of the cloud to be simplified. KNN strictly depends on the density of the cloud: a very dense cloud, as fountain-P11, requires a very high K.


\subsection{Incremental Point Cloud Simplification}
The main goal of point simplification is to improve the trade-off between accuracy and speed, therefore a typical target application is to improve the performance of an incremental mapping algorithm as~\cite{piazza2018real}, which runs in real time on a CPU by using Structure from Motion points, but is not able to handle very dense point clouds still maintaining realtime performance.
To test the effectiveness of our proposal in this scenario, we added the dense point estimation and our Probabilistic Simplification module to the algorithm proposed in~\cite{piazza2018real}, which is the  incremental version of the pipeline described in \secref{delaunay}. 

We tested  the sequences 
$95$ and $104$ of the  KITTI dataset~\cite{geiger_et_al12}, and we compute the model accuracy in the same way. The semantic segmentation are computed using~\cite{zhou2017scene} pre-trained over ADE20K dataset (only 11 among the 150 classes have been used).

In \tabref{incremental_kitti} we compare the incremental mapping performance from dense point clouds with or without the proposed simplification. The accuracy is measured with the RMSE adopted in the previous section, while the speed is the total number of frames of the sequence over the seconds spent to build all the mesh incrementally.
The results show how the simplification method is able to keep a similar accuracy while improving the frame rate. 

Speed-up depends on the scene composition. Sequence $104$ contains many elements which are not addressed by the simplification procedure, \eg, trees, vegetation and cars, therefore the frame rate increase is limited. 
Instead, 
sequence $95$ is an heterogeneous scenes, therefore the frame rate increases of more than $50\%$.
\begin{table}[t]
    \caption{Incremental Mapping results. }
    \label{tab:incremental_kitti}

    \centering
    \begin{tabular}{c|cc|cc}
    &
    \multicolumn{2}{c|}{Original} &
    \multicolumn{2}{c}{Incremental Simplification} \\
    Sequence & Accuracy & Frame rate & Accuracy & Frame rate \\
             & RMSE(m)  & (f/s)      & RMSE(m)  & (f/s)      \\
    \hline
    95       & \textbf{0.7135}   & 1.0082     & 0.7173   & \textbf{1.5434}     \\
    104      & \textbf{0.7775}   & 0.4708     & 0.7846   & \textbf{0.6397}     \\
    \end{tabular}
\end{table}

\section{Conclusions}
In this paper, we leverage on semantics to simplify redundant regions of a dense point cloud prior to build a 3D mesh upon to its points.
We fuse these points into a Delaunay triangulation which easily adapts to the point cloud even if we decrease the density of some regions, and we extract a visibility consistent mesh.
We designed four methods which provide different alternatives to decimate the point cloud, halving its size without relevant loss in accuracy. 
Among these methods the probabilistic approach in general is easier to tune and proved to be able to identify a good trade off between simplification and map accuracy.
We have also shown how a state-of-the-art incremental mapping method can benefit from the proposed semantic point simplification to process less redundant data and to consequently speed-up the mapping.
As future work, we plan to evaluate viable energy efficient optimizations~\cite{Zoni:2015:MDP:2828988.2751561} for the proposed simplification procedure and we plan to design a flexible method able to deal with every semantic class, \eg, by identifying a per class simplification probability.


\bibliographystyle{splncs}
\bibliography{main}
\end{document}